\documentclass{article} 
\usepackage{iclr2025_conference,times}

\usepackage{diagbox}
\usepackage{hyperref}
\usepackage{arydshln}
\usepackage{array}
\usepackage{url}
\usepackage{amsmath}
\usepackage{amssymb}
\usepackage{bbm}
\usepackage{graphicx}
\usepackage{caption}   
\usepackage{makecell}
\usepackage{amsfonts}
\usepackage{bm}
\usepackage{multirow}

\def\eg{\emph{e.g.}} 

\def\ie{\emph{i.e.}}

\title{Toward Generalizing Visual Brain Decoding to Unseen Subjects}

\author{
    Xiangtao Kong $^{1}$\thanks{Equal contributions.},
    Kexin Huang $^{1}$\footnotemark[1] , 
    Ping Li $^{1}$\thanks{Corresponding authors.},
    Lei Zhang $^{1}$\footnotemark[2]  \\
$^{1}$The Hong Kong Polytechnic University \\
\texttt{\{xiangtao.kong, kexin0.huang\}@connect.polyu.hk,}\\
\texttt{ping2.li@polyu.edu.hk, cslzhang@comp.polyu.edu.hk}
}

\iclrfinalcopy 
\begin{document}

\maketitle

\begin{abstract}

Visual brain decoding aims to decode visual information from human brain activities. Despite the great progress, one critical limitation of current brain decoding research lies in the lack of generalization capability to unseen subjects. Prior works typically focus on decoding brain activity of individuals based on the observation that different subjects exhibit different brain activities, while it remains unclear whether brain decoding can be generalized to unseen subjects. This study aims to answer this question. We first consolidate an image-fMRI dataset consisting of stimulus-image and fMRI-response pairs, involving 177 subjects in the movie-viewing task of the Human Connectome Project (HCP). This dataset allows us to investigate the brain decoding performance with the increase of participants. We then present a learning paradigm that applies uniform processing across all subjects, instead of employing different network heads or tokenizers for individuals as in previous methods, which can accommodate a large number of subjects to explore the generalization capability across different subjects. A series of experiments are conducted and we have the following findings. First, the network exhibits clear generalization capabilities with the increase of training subjects. Second, the generalization capability is common to popular network architectures (MLP, CNN and Transformer). Third, the generalization performance is affected by the similarity between subjects. Our findings reveal the inherent similarities in brain activities across individuals. With the emerging of larger and more comprehensive datasets, it is possible to train a brain decoding foundation model in the future. Codes and models can be found at \href{https://github.com/Xiangtaokong/TGBD}{https://github.com/Xiangtaokong/TGBD}.


\end{abstract}

\section{Introduction}
\label{sec:Introduction}

Visual brain decoding \citep{kay2008identifying,kamitani2005decoding,naselaris2011encoding} aims to decode visual information from human brain activities, including tasks of brain-image classification \citep{kaur2019automated,CLIP-MUSED}, retrieval \citep{scotti2024reconstructing,umbrae} and reconstruction \citep{takagi2023high,braindiffuser,Throughtheireyes,scotti2024reconstructing}, and so on.
It involves analyzing neural patterns collected via brain imaging techniques like functional magnetic resonance imaging (fMRI) \citep{schirrmeister2017deep, benchetrit2023brain, kamitani2005decoding} or electroencephalography (EEG) \citep{schirrmeister2017deep, vallabhaneni2021deep} to infer the visual information received by the participants.
Among them, fMRI is favored by researchers because of its more informative depiction of the whole brain activity, which has resulted in a number of important decoding works \citep{nsd,takagi2023high,scotti2024reconstructing} with the help of deep learning techniques.

A major limitation of current brain decoding research, however, lies in the lack of generalization capability to unseen subjects. 
That is, the trained decoding models can hardly be applied to new, unseen individuals. 
Such a limitation can be owed to two reasons. 
First, there are individual differences of the brain activities across subjects \citep{haxby2020hyperalignment}. 
Therefore, it is assumed that brain decoding cannot be generalized and hence researchers are focused on developing subject-specific models.
Second, commonly used brain visual decoding datasets are built upon only a small number of participants. 
For example, the Natural Scenes Dataset (NSD) dataset \citep{nsd} includes only 8 subjects. 
Most NSD-based studies \citep{kaur2019automated,scotti2024reconstructing} employ only 4 of the 8 subjects, and use the NSDGeneral data, which contain only the manually mapped brain region, rather than the entire brain data (See more detailed discussions in Sec. \ref{sec: dataset}).
Even those studies attempting to leverage multiple subjects are typically limited to less than 10 participants, and their networks are designed to handle only a small number of individuals. 
For instance, MindEye2 \citep{scotti2024mindeye2} and UMBRAE \citep{umbrae} use separate heads or tokenizers for different subjects. Therefore, the model becomes increasingly complex as the increase of subject number, which is hard to scale up to a larger number of subjects.

\begin{figure}[t]
\begin{center}
\includegraphics[width=0.88\linewidth]{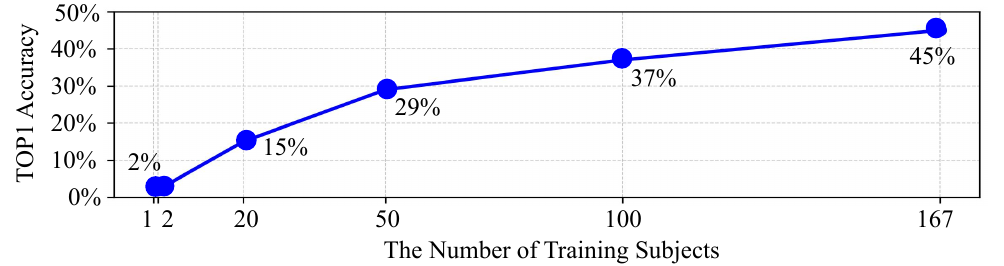}
\end{center}
\vspace{-0.4cm}
\caption{The performance on unseen subjects with the increase of the number of training subjects.}
\vspace{-0.8cm}
\label{fig:line167}
\end{figure}

In this work, we aim to address this limitation and answer the question that whether brain decoding can be generalized to unseen subjects.
To this end, we first consolidated an image-fMRI dataset, which consists of pairs of the stimulus image and the corresponding brain fMRI response. We build this dataset using the data from the Human Connectome Project (HCP) \citep{HCP}, which contains human brain neuroimages for various tasks. 
Among them, 177 subjects participated in the movie-viewing task, which provides the largest number of subjects available for extracting image-fMRI pairs for visual decoding study. 
In total, we collected 3,127 data pairs from 4 films watched by the 177 subjects. 
Compared to the commonly used datasets like NSD (8 subjects) \citep{nsd} and BOLD5000 (4 subjects) \citep{chang2019bold5000}, this dataset enables us to explore brain decoding performance with a much larger number of subjects. 
We consequently propose a new learning paradigm.
Following MindEye1 \citep{scotti2024reconstructing}, we use CLIP to encode the images, and employ a brain decoding network to map brain activities (characterized by fMRI voxels) into the same CLIP space by contrastive learning. 
To handle the varying fMRI voxel sizes across subjects, we simply normalize them to a common size through upsampling.
Unlike previous methods that rely on specially designed input heads or subject-specific tokenizers, our paradigm uses the same processing for all subjects so that it can handle a large number of subjects without increasing the model complexity and parameters.

We perform experiments on the fundamental retrieval task, which reflects well the capabilities of decoding models. (Our method can be extended to reconstruction or grounding tasks with additional modules such as Stable Diffusion \citep{SD}.)
Through detailed experimentation, we uncover several important and intriguing findings. 
First, as shown in Fig. \ref{fig:line167}, the network demonstrates clear generalization ability as the number of training subjects increases, with top-1 accuracy rising from 2\% (1 training subject) to 45\% (167 training subjects) on unseen subjects (100 image-fMRI pairs). The accuracy can be further improved to 50\% with additional training strategies.
Second, the generalization capability holds for different network architectures.
Using MLP, CNN and Transformer as the backbone, we achieve top-1 accuracies of 45\%, 42\%, and 34\%, respectively, with 167 training subjects.
Third, the generalization performance is influenced by subject similarity.
We observe a bias when training on distinct groups such as gender, which represents one of the most easily identifiable categories of sample similarity. 
The model trained on 50 males achieves 36\% top-1 accuracy on an unseen male subject, while the model trained on 50 females only obtains 27\% top-1 accuracy on this test.
Therefore, to explore further, we design an algorithm to calculate the similarity of fMRI responses among 167 individuals and train two models on the 20 most similar and 20 least similar subjects.
The models achieve 21\% and 2\% top-1 accuracy, respectively, on an unseen subject, indicating the degree of similarity across subjects greatly affects the generalization performance.
Our findings reveal that human brain activities share similarities, which is worth of further exploration. It may be possible to train a large foundation model for brain decoding as bigger and more comprehensive datasets emerge.



\section{Related Work}
\label{sec:Related Work}


\paragraph{Visual Brain Decoding.}
With the advancement of deep learning \citep{CLIP,ResNet,Transformer,SD} and the emergence of high-quality fMRI datasets \citep{nsd,chang2019bold5000,HCP}, many brain decoding methods with promising performance have been proposed.
\cite{takagi2023high} utilized a latent diffusion model, specifically Stable Diffusion \citep{ho2020denoising,sohl2015deep}, to reconstruct high-resolution images from fMRI data, preserving semantic fidelity without requiring additional training or fine-tuning.
Brain-Diffuser \citep{braindiffuser} improves the reconstruction process by first reconstructing basic image properties from fMRI signals and then refining the images using a latent diffusion model conditioned on multimodal features.
Moreover, MindEye~\citep{scotti2024reconstructing} encodes images using CLIP and then maps the corresponding fMRI data to the CLIP feature space, enabling strong image retrieval or reconstruction performance.
However, most existing methods focus on decoding stimuli for individual subjects.
While effective for individual decoding, they lack the generalization capability to new, unseen subjects.
%

\vspace{-2mm}

\paragraph{Visual Brain Decoding on Multiple Subjects.}
Some brain decoding methods have been developed to leverage multiple subjects, which can be categorized into two categories based on their objectives: (1) using multiple subjects to enhance subject-specific models, and (2) developing models that handle multiple subjects directly.
For the first category, a straightforward way is to pre-train models on multiple subjects and then fine-tune them for individual subjects \citep{scotti2024mindeye2, jiang2024mindshot, qian2023fmri,Throughtheireyes}.
For example, MindEye2~\citep{scotti2024mindeye2} pre-trains the model on 7 subjects from the NSD dataset~\citep{nsd} and fine-tunes it on a different subject, using only 1/40 of the original data while achieving similar performance.
The second category of methods aim to train a model with multiple subjects so that its performance on each subject (included in the training set) surpasses the models trained on each single subject. 
CLIP-MUSED \citep{CLIP-MUSED} and UMBRAE \citep{umbrae} are methods of this kind. 
However, most methods in this category still require separate heads or tokenizers for each subject. As the number of subjects increases, their training costs and model parameters grow linearly, making this approach impractical for larger subject pools.
These multi-subject methods generally involve a limited number of subjects (less than 10), which are not sufficient enough for exploration. 
More importantly, while these methods demonstrate that certain information can be shared across subjects, they cannot generalize to unseen subjects.

\vspace{-2mm}
\paragraph{Subjects Alignment.}
Some methods have been proposed to align new subjects to pre-trained models, known as subject alignment, to handle unseen subjects. 
Based on the alignment approach, these methods can be categorized into anatomical alignment \citep{jenkinson2002improved} and functional alignment \citep{haxby2011common,lorbert2012kernel,xu2012regularized,chen2015reduced}, among others.
In visual brain decoding, the mainstream methods fall into functional alignment, which directly aligns the neural activity patterns across different subjects.
For instance, \cite{Throughtheireyes} used 1,000 common images viewed by 8 subjects from the NSD dataset to train an alignment model that maps other subjects to Subject 1.
During inference, the brain signals of other subjects are converted into the format of Subject 1 and fed into the model trained on Subject 1. 
This approach can process new subjects with the model of existing subjects at a lower cost, yet it requires shared data for alignment.
In this work, we aim to achieve model generalization without such alignment.

\section{Methods}
\label{sec:Methods}

In this section, we first describe how we consolidate the dataset for exploring generalizable visual brain decoding in Sec. \ref{sec: dataset}. 
Then, we describe the proposed learning paradigm in Sec. \ref{sec:Learning Paradigm}.
Finally, in Sec. \ref{sec:Calculating similarity based on rank} we outline how we calculate the subject similarity.

\subsection{Dataset Consolidation}
\label{sec: dataset}

\begin{table}[t]

\setlength\tabcolsep{3pt}
\renewcommand\arraystretch{1.4}
\caption{Summary of commonly used visual brain decoding datasets.}
\label{table:dataset}
\vspace{-0.1cm}
\centering
\scalebox{0.82}{
\begin{tabular}{>{\centering\arraybackslash}m{5.3cm}>{\centering\arraybackslash}m{2.2cm}>{\centering\arraybackslash}m{1.0cm}>{\centering\arraybackslash}m{1.2cm}>{\centering\arraybackslash}m{6.2cm}}
\hline
Dataset           & Task                & Scanner & Subjects & Works Based on This Dataset                          \\ \hline
BOLD5000 \citep{chang2019bold5000}         & image-viewing       & 3T         & 4        &  \cite{chen2023seeing,prince2022improving,sexton2022reassessing}                 \\ \cdashline{1-5}
GOD \citep{horikawa2017generic}&image-viewing& 3T & 5&\cite{chen2023seeing,du2023decoding}\\ \cdashline{1-5}
NSD \citep{nsd}             & image-viewing       & 7T         & 8        & MindEye1\&2 \citep{scotti2024reconstructing,scotti2024mindeye2};  \cite{gu2022decoding,Throughtheireyes,qian2023fmri,han2024mindformer} \\ \hline
Raiders \citep{haxby2011common}           & movie-viewing & 3T         & 21        &  \cite{chen2015reduced,shvartsman2018matrix}                                                   \\ \cdashline{1-5}

Forrest Gump \citep{hanke2014high} & movie-viewing & 7T         & 20        &   \cite{chen2015reduced,wagner2022fairly,huang2022learning}                                                   \\ \cdashline{1-5}
Budapest \citep{visconti2020fmri}& movie-viewing & 3T & 25 & \cite{visconti2021shared,busch2021hybrid} \\ \cdashline{1-5}
HCP \citep{HCP}              & movie-viewing & 7T         & 177      & \cite{CLIP-MUSED,lu2024animate}                                                     \\
\hline
\end{tabular}
}

\end{table}

Most previous studies \citep{scotti2024reconstructing,scotti2024mindeye2,umbrae,CLIP-MUSED} are conducted on datasets with fewer than ten participants, which cannot be used to study whether visual decoding can be generalizable. 
Therefore, to explore the generalization capabilities of brain decoding models, the first step is to collect a dataset with a larger number of subjects. 
However, as shown in Tab. \ref{table:dataset}, current publicly available image-viewing datasets are limited in size. For example, the NSD dataset involves only 8 subjects \citep{nsd} and BOLD5000 involves only 4 subjects \citep{chang2019bold5000}. 
This is mainly due to the high costs, time demands, and challenges in keeping participants engaged during the long fMRI scanning sessions.
Furthermore, these existing datasets are hard to be combined due to their significant differences in scanning equipment, resolution, and post-processing methods.
Even combined, the total number of subjects remains very small.

We then turn to the movie-viewing task, which provides continuous, causally-related visual inputs over a short period. 
Compared to the image-viewing task, movie-viewing can yield much more data pairs in the same time-frame while keeping participants more engaged. Therefore, movie-viewing experiments can involve more subjects, as shown in Tab. \ref{table:dataset}.
Actually, some works have been proposed to extract video frames for visual brain retrieval \citep{schneider2023learnable} and classification \citep{CLIP-MUSED}.
Therefore, we propose to extract image-fMRI pairs from the movie-viewing task to build our dataset.
Specifically, we choose the movie-viewing task in the HCP dataset, which involves 177 participants, making it the largest movie-viewing dataset available for visual decoding research.
In data collection, the participants watched four audiovisual films, and the fMRI responses were captured with a repetition time (TR) of 1 second using a high-resolution 7T scanner.
The volumetric images are registered to 1.6mm MNI space, with dimensions of 113 × 136 × 113 per TR.

\begin{figure}[t]
\begin{minipage}{0.7\textwidth}
    \centering
    \includegraphics[width=\linewidth]{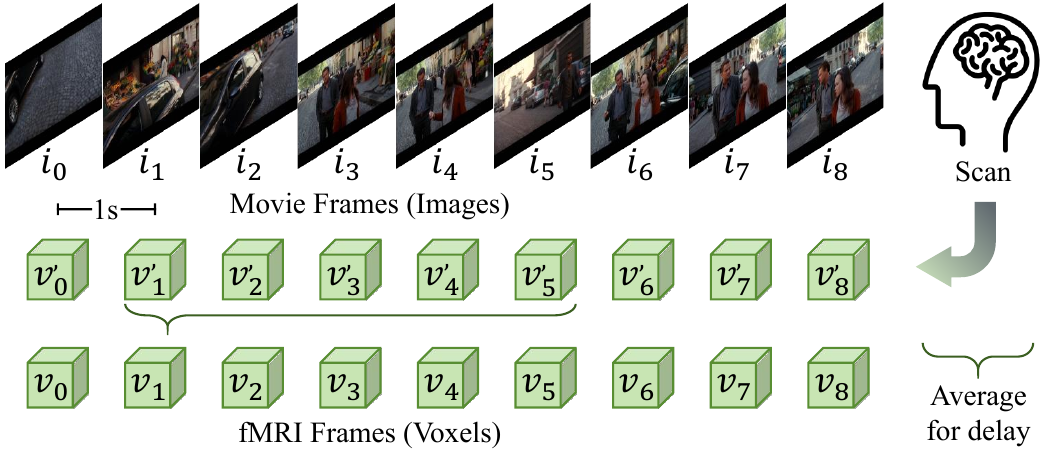} 
\end{minipage}%
\begin{minipage}{0.3\textwidth}
    \centering
    \captionof{figure}{Dataset reconstruction from the HCP data. We extract the last frame $i$ in each second of the movie clip as the stimulus image, and average the fMRI voxels in the subsequent 4 seconds (due to hemodynamic delay) as the corresponding neural response $v$ to obtain image-fMRI pairs.}  
    \label{fig:hcpdata}
\end{minipage}
\vspace{-0.8cm}
\end{figure}

As shown in Fig. \ref{fig:hcpdata}, to extract the corresponding image-fMRI pairs, we extract the last frame $i$ of each second of the film as the stimulus image, whose corresponding fMRI response voxel is denoted as $v^,_i$.
Following the 4-second hemodynamic delay suggested by \cite{khosla2020}, we average the fMRI signals from the subsequent four seconds to represent the neural response to each stimulus image (\eg, average $v^,_1$ - $v^,_5$ to obtain $v_1$ for $i_1$), resulting in a total of 3,127 image-fMRI pairs for each subject.
Finally, the reconstructed dataset includes 177 subjects with 177 × 3,127 image-fMRI pairs.
By leveraging this dataset, we can explore the generalization capabilities of brain decoding models across a broader population.
Our experiments in Sec. \ref{sec:Discovering generalization} demonstrate that generalization will emerge when a sufficient number of subjects are involved in training. 
This dataset provides a valuable foundation for researchers to investigate the behaviours of brain decoding models.

\subsection{Learning Paradigm}
\label{sec:Learning Paradigm}

Prior studies are typically focused on decoding brain activity of individuals, while little work has been done on exploring the model generalization capability to unseen subjects. 
With our consolidated dataset in Sec. \ref{sec: dataset}, we propose a learning paradigm to investigate the generalizability of visual brain decoding based on three core principles:
(1) utilization of whole-brain data;
(2) simple and flexible pipeline;
(3) applicability to a large number of diverse subjects.

\paragraph{Utilization of the Whole-Brain Data.}


\begin{figure}[t]
\begin{center}
\includegraphics[width=0.8\linewidth]{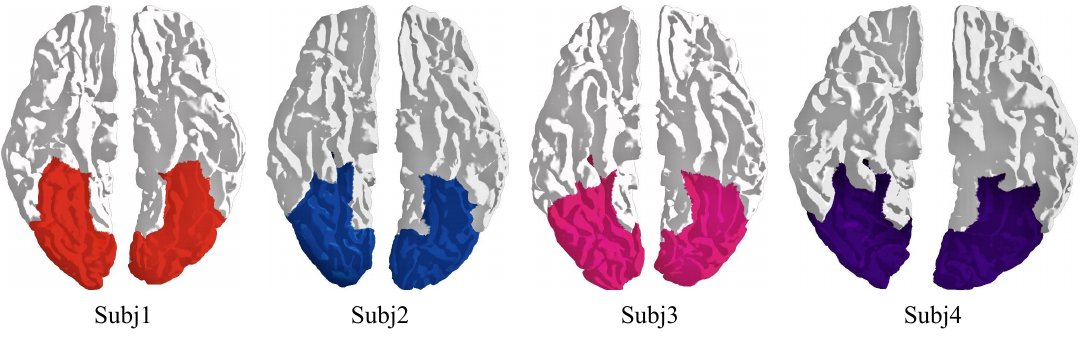}
\end{center}
\vspace{-0.4cm}
\caption{The visualization of scanned brain data in NSD dataset. The highlighted regions indicate the manually labeled NSDGeneral data. Compared to the whole brain, the NSDGeneral regions show significant variations across different subjects.}
\label{fig:nsdgeneral}
\end{figure}

As shown in Tab. \ref{table:dataset}, most recent visual brain decoding studies rely on the NSD dataset, which provides two types of training data: NSDGeneral data and whole-brain data. 
As illustrated in Fig. \ref{fig:nsdgeneral}, the whole-brain data contain the fMRI voxels (about 800K elements) of the entire brain, while the NSDGeneral data comprise 1D vectors (flattened voxels) of only 10k–20k elements, which are manually labeled as vision-related brain regions, called NSDGeneral regions (see the highlighted areas).
Since NSDGeneral data are directly related to brain regions in charge of visual processing, they often result in better visual decoding performance in the studies focused on single subjects.
However, for research investigating the generalization capability across multiple subjects, we believe whole-brain data are more appropriate. 
First, the NSDGeneral data require manual segmentation, and hence they are difficult to scale across a large number of subjects, limiting their suitability for generalization studies.
Note that most datasets, such as the HCP dataset, only provide whole-brain data.
Second, as can be seen in Fig. \ref{fig:nsdgeneral}, the manually labeled NSDGeneral regions show significant variations across different subjects, while the whole brain data show much less variations in shape.
It is thus more difficult to train a common model to multiple subjects using the NSDGeneral data than the whole-brain data (See Sec. \ref{sec:Results on NSD dataset}).
Third, the NSDGeneral data exclude other brain regions, such as those in charge of memory or contextual understanding. 
Ignoring those regions may prevent a more comprehensive decoding of brain activities \citep{CLIP-MUSED}.
Therefore, we advocate for using whole-brain data in the study of on model generalization, rather than data limited to specific brain regions.

\paragraph{Simple and Flexible Pipeline.}

\begin{figure}[t]
\begin{center}
\includegraphics[width=1\linewidth]{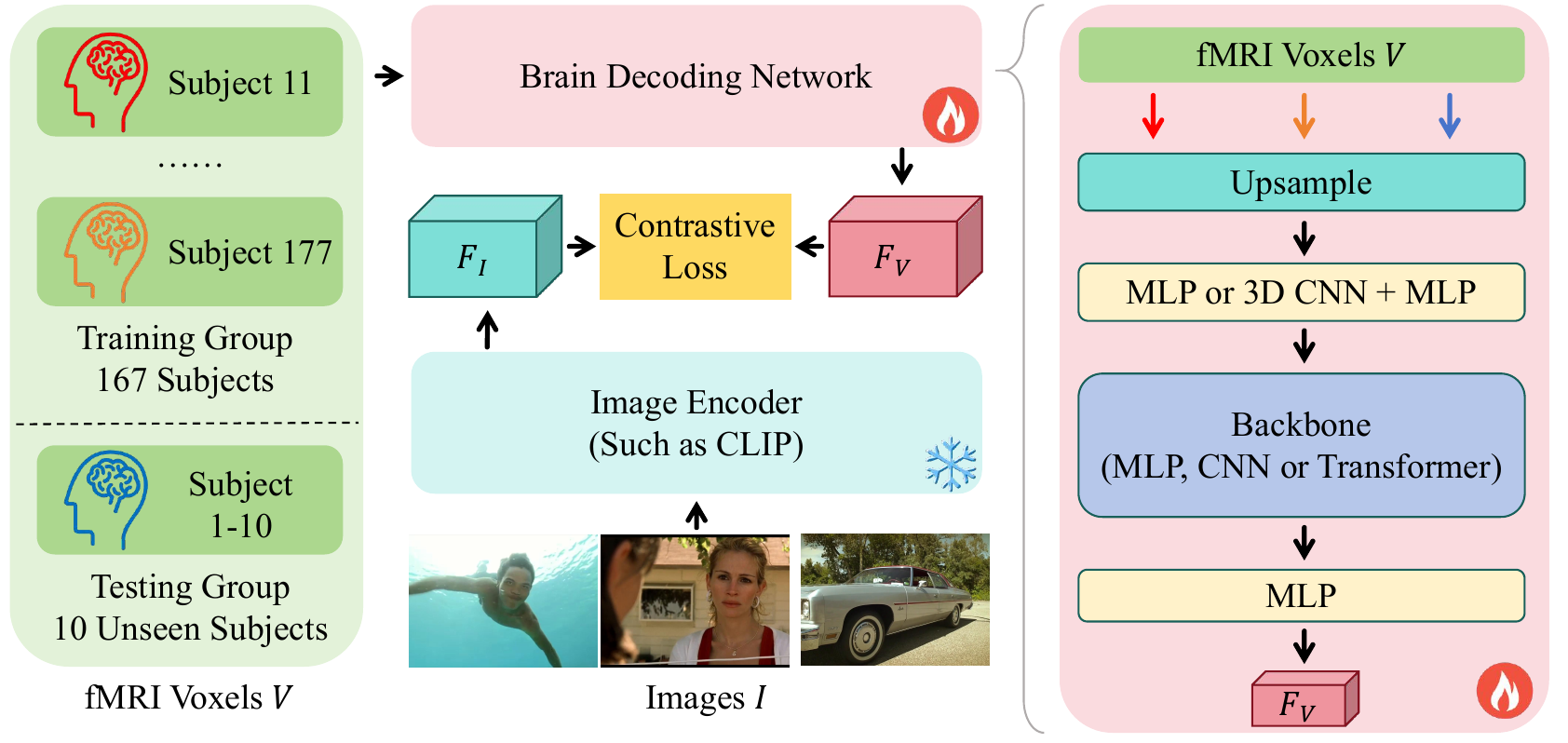}
\end{center}
\vspace{-0.2cm}
\caption{The overview of our learning pipeline (left) and visual brain decoding network (right).}
\label{fig:pipeline}
\vspace{-0.2cm}
\end{figure}

A simple and flexible learning pipeline is preferred to verify whether generalizability is a fundamental property of visual brain decoding, minimizing the factors brought by complex network designs. 
Our learning pipeline is shown in the left part of Fig. \ref{fig:pipeline}. The core idea is to project the paired stimulus-image $I$ and fMRI-voxel $V$ into the same feature space, where they could be as similar as possible.
Following \citep{scotti2024reconstructing, umbrae}, we use the CLIP ViT-L/14 model to encode the images into features $F_I$, while the visual brain decoding network is trained to map fMRI-voxels to $F_V$ in the same feature space.
The feature size of the CLIP embedding space is $257\times1024$, which retains detailed image information compared to the high-level semantic content of the final CLS token in CLIP.
Contrastive learning is employed to align $F_I$ with $F_V$ using the CLIP Loss:
\begin{equation}
\mathcal{L} = \frac{1}{2N} \left( \sum_{i=1}^{N} -\log \frac{\exp(\text{sim}(F_I^i, F_V^i) / \tau)}{\sum_{j=1}^{N} \exp(\text{sim}(F_I^i, F_V^j) / \tau)} + \sum_{i=1}^{N} -\log \frac{\exp(\text{sim}(F_V^i, F_I^i) / \tau)}{\sum_{j=1}^{N} \exp(\text{sim}(F_V^i, F_I^j) / \tau)} \right),
\end{equation}
where $F_I^i$ and $F_V^i$ are the embeddings of the $i^{th}$ image and fMRI voxel, $\tau$ is the temperature parameter, and $\text{sim}(x, y)$ represents the cosine similarity between $x$ and $y$:
\begin{equation}
\label{equation:sim}
\text{sim}(x, y) = \frac{x \cdot y}{\|x\| \|y\|}.
\end{equation}
During inference, retrieval is performed by calculating the cosine similarity and taking the most similar pairs.
%
We use MLP as the decoding network in most of our experiments, while the network architecture can be changed to CNN and Transformer (See Sec. \ref{sec:Verifying generalization}), and the model performance can be further improved with additional strategies (See Sec. \ref{sec:Training strategy}).

\paragraph{Applicability to Many Diverse Subjects.}

Previous studies are mostly focused on a small number of participants, and they use separate heads or tokenizers for different subjects to improve performance. 
While being effective in small scale studies, these approaches become impractical and cannot scale up as the number of subjects increases.
To explore model generalizability, we employ the same decoding network (see the right part of Fig. \ref{fig:pipeline}) to accommodate a large number of subjects without requiring specific adaptations for each individual.
Due to the structural differences in brain anatomy, the size of fMRI voxels, even for the same brain activity, can vary across subjects, which cannot be directly batched for network training.
To solve this issue, we apply simple upsampling to resize the voxels to a standardized larger size.
This method is simple and straightforward and can be done in the processing of the dataset.
%
Experimental results show that this unified approach does not compromise performance in whole-brain decoding and can even enhance performance across multiple subjects (See Sec. \ref{sec:Results on NSD dataset}).
As shown in the right of Fig. \ref{fig:pipeline}, our final network design involves an upsample layer to normalize voxel sizes for all subjects, followed by feeding the data into a unified network without requiring any subject-specific adaptations.

\subsection{Generalization Performance vs. Subject Similarity}
\label{sec:Calculating similarity based on rank}

During experiments, we notice some performance biases of models trained on different gender groups, which represent one of the most easily identifiable similarity categories.
It inspires us to hypothesize that the degree of similarity among subjects might impact the model generalization performance. 
To test this hypothesis, we need to identify which subjects are more similar to the given subjects.
For a target subject $S_t$, given a set of images $I$ viewed by $N$ different subjects $S_N$, for each image $i$, we can calculate the cosine similarity (refer to Eq. \ref{equation:sim}) between the fMRI voxel $v_{i,S_t}$ of target subject and the voxel $v_{i,S_n}$ of subject $S_n \in S_N$ as follow: $Sim\_score_{i,S_t,S_n}=sim(v_{i,S_t},v_{i,S_n})$.
The similarity score reflects the likeness between the two subjects ($S_t$ and $S_n$) based on a given image $i$. The overall similarity score of the two subjects can be obtained by averaging over all images $I$. However, the outlier images can make the averaged score less robust. 
Therefore, we use a rank-based method. For each image $i$, we calculate $Sim\_score_{i,S_t,S_n}$ and rank the $N$ scores from highest to lowest.
Then, we select the top 10 subjects based on their ranks and award them one rank credit.
After repeating this process for all images, subjects with higher total rank credits are considered more similar to the target subject $S_t$.
The process to calculate the rank credit of $S_n$ for $S_t$ can be formulated as:
\begin{equation}
    \text{Rank\_Credit}(S_t,S_n) = \sum\nolimits_{i=1}^{I} \mathbbm{1} (S_n \in \text{top\_10\_rank}(Sim\_score_{i,S_t,S_j} \text{ for } j=1, 2 \ldots, N)),
\end{equation}
%
where $\mathbbm{1}(\cdot)$ is an indication function that assigns 1 if $S_n$ is among the top 10 most similar subjects to $S_t$ based on image $i$, otherwise 0.
Finally, we use models trained on both similar and dissimilar subjects to explore how subject similarity influences the generalization performance. The results are shown in Sec. \ref{sec:Exploring generalization}.

\section{Experiments}
\label{sec:Experiments}

\subsection{Implementation Details}
\label{sec:Implementation Details}


We implement all models using PyTorch~\citep{pytorch}.
Except specifically indicated, we employ MLP and 3D CNN as the backbone for feature extraction when using whole-brain data.
%
The detailed network structure can be found in the \textbf{appendix}.
During training, we employ the CLIP loss~\citep{CLIP} and the AdamW optimizer~\citep{adamw} to optimize the models ($\beta_1$ = 0.9, $\beta_2$ = 0.999).
We set the batch size to 300, and apply the OneCycleLR strategy with a warm-up phase to adjust the learning rate, with a maximum learning rate of $1\times10^{-4}$.
The HCP dataset we consolidated includes 177 subjects, each subject having 3,127 image-fMRI pairs.
We randomly choose 100 images and the corresponding fMRI voxels as the test pairs, and use the rest as the training pairs.
Note that the test pairs of all subjects are from the same 100 images.
Subjs 1-10 are designated as unseen subjects, with the remaining 167 subjects as seen subjects.

In our experiments, several models will be trained on different numbers of subjects. For convenience of expression, we define one training epoch based on the number of image-fMRI pairs of a single subject; that is, one epoch contains 3,027 image-fMRI pairs.
The numbers of epochs to train models on 1, 2, 20, 50, 100, and 167 seen subjects are 200, 200, 400, 600, 800, and 1,000, respectively.
For the experiment on the NSD dataset, which includes 8 subjects, we follow the standard train/test split with 1,000 test images \citep{nsd}, and select Subj 2 and Subj 5 as unseen subjects.
We train the models with 1 and 6 seen subjects for 120 and 360 epochs, respectively, and each epoch includes 9,000 image-fMRI pairs.

\subsection{Main Results on Generalization Performance}
\label{sec:Discovering generalization}



\begin{table}[t]
\setlength\tabcolsep{3pt}
\renewcommand\arraystretch{1.3}
\caption{Results of models trained on our consolidated HCP dataset with different number of training subjects. TOP1 Acc. and TOP3 Acc. are averaged over the unseen subjects (Subjs 1-10). TOP1 Acc. (seen) is averaged over all seen subjects participated in training.}
\label{table:main}
\centering
\scalebox{1}{
\begin{tabular}{ccccc}
\hline
Training Subjects & No. of Training Subjects & TOP1 Acc.  & TOP3 Acc. & TOP1 Acc. (seen)  \\ \hline
Subj 11 & 1 & 2\%  & 5\% & 79\%   \\
Subjs 11-12  & 2 & 2\%  & 6\% & 84\%  \\
Subjs 11-30  & 20 & 15\%  & 29\% & 83\%  \\
Subjs 11-60  & 50 & 29\%  & 43\% & 83\%  \\
Subjs 11-110  & 100 & 37\% & 52\% & 82\%  \\
Subjs 11-177  & 167 & 45\%  & 61\% & 82\%  \\ \hline

\end{tabular}
}
\end{table}

%
As describe in Sec. \ref{sec:Implementation Details}, we train models on 1, 2, 20, 50, 100 and 167 subjects and evaluate them on 10 unseen subjects (Subjs 1-10).
The results are shown in Tab. \ref{table:main}. We can clearly see that as the number of training subjects increases, the model's generalization capability on unseen subjects improves. 
When only one or two subjects are used in training, the generalization capability is weak. When the number of training subjects reaches 167, the TOP1 and TOP3 accuracies improve to 45\% and 61\%, respectively. 
Considering that our test set contains 100 image-fMRI pairs, such a generalization performance is highly encouraging.
Fig. \ref{fig:line167} plots the curve of TOP1 accuracy vs. the number of training subjects. We observe that the generalization performance continues to improve steadily. Even with 167 subjects, it does not reach a plateau.
This suggests that the models hold potential for further improvement if more subjects can be introduced for training.

We also provide in Tab. \ref{table:main} the results of our model on seen subjects for reference.
The test pairs are from the subjects involved in the training process. We can see that the TOP1 accuracies are very close for different number of training subjects.
This is reasonable because the testing data share similar distribution with the training data, no matter 1 or 167 subjects are involved, and the network is able to fit the distribution via sufficient training. 

\subsection{The Generalization Performance with Different Backbones}
\label{sec:Verifying generalization}


\begin{table}[t]
\setlength\tabcolsep{3pt}
\renewcommand\arraystretch{1.3}
\centering
\begin{minipage}{0.6\textwidth}
    \centering
    \scalebox{0.9}{
    \begin{tabular}{cccc}
\hline
Backbones & TOP1 Acc.  & TOP3 Acc. & TOP1 Acc. (seen) \\ \hline
MLP & 45\%  & 61\% & 82\%  \\
1D CNN & 42\%  & 58\% & 80\%  \\
3D CNN & 40\%  & 57\% & 80\%  \\
Transformer & 34\%  & 52\% & 77\%  \\
 \hline
\end{tabular} 
}
\end{minipage}%
\begin{minipage}{0.4\textwidth}
    \centering
\vspace{6pt}
\caption{Results of models with different backbones trained on 167 subjects. TOP1 Acc. and TOP3 Acc. are averaged over ten unseen subjects (Subjs 1-10). TOP1 Acc. (seen) is averaged over all seen subjects.}
\label{table:backbone}

\end{minipage}
\vspace{-0.3cm}
\end{table}

In Sec. \ref{sec:Discovering generalization}, we used MLP as the backbone and validated the generalization capability of our models trained on more subjects. Here, we validate whether this conclusion holds for other popular network architectures, such as CNN and Transformer networks. 
The details of the employed network architectures can be found in the \textbf{appendix}.
The results are shown in Tab. \ref{table:backbone}, which demonstrates that while there are some differences in performance, the generalization capability is consistently achieved across different network architectures.
Specifically, the MLP achieves the best generalization performance, with 45\% TOP1 accuracy on unseen subjects and 82\% on seen subjects, followed closely by 1D CNN and 3D CNN, which yield comparable results. 
The Transformer network exhibits the lowest performance, which may be attributed to the fact that Transformer typically needs larger training datasets to exhibit its superiority, whereas our current dataset for visual brain decoding is relatively small, making CNNs and MLPs more effective in this case.

\subsection{Generalization vs. Subject Similarity}
\label{sec:Exploring generalization}

From the experiments in previous sections, we have seen that the network could exhibit obvious generalization capability when enough subjects are used in training.
%
During our experiments, interestingly, we notice some performance biases of models trained on different gender groups.
Gender is one of the most commonly observed characteristics of sample similarity, which inspires us that the similarity among subjects might impact the model generalization performance.
To be specific, we train three models using data from 50 male subjects, 50 female subjects, and a mixed group of 25 male and 25 female subjects, respectively.
Then, we evaluate these models on unseen male Subj 1 and female Subj 2, and the results are shown in Tab. \ref{table:similar}.
One can see that the model trained on male subjects achieves the best retrieval performance on the unseen male subject (Subj 1) with 36\% TOP1 accuracy and 60\% TOP3 accuracy, but it performs the worst on the unseen female subject (Subj 2) with 25\% TOP1 accuracy and 37\% TOP3 accuracy. 
In contrast, the model trained on female subjects shows the opposite behaviour, performing better on Subj 2 and worse on Subj 1.
Meanwhile, the model trained on the mixed group always obtain the intermediate result on the unseen subjects.
Such results suggest that the generalization capability is related to the similarity among subjects.

Therefore, we further explore this phenomenon by finding similar and dissimilar subjects to Subj 1 and Subj 2.
Utilizing the method described in Sec. \ref{sec:Calculating similarity based on rank}, we identify 20 most similar subjects and 20 least similar subjects to Subj 1, as well as 20 most similar and least similar subjects to Subj 2.
As shown in Tab. \ref{table:similar}, the model trained on the 20 subjects most similar to Subj 1 achieves the best performance on Subj 1, with a TOP1 accuracy of 21\% and a TOP3 accuracy of 36\%. 
In contrast, the model trained on the 20 most dissimilar subjects performs significantly worse, with a TOP1 accuracy of 8\% and a TOP3 accuracy of 14\%.
A similar trend can be observed for Subj 2, where the model trained on the 20 most similar / dissimilar subjects achieves the highest / lowest performance.
Additionally, the model trained on Subjs 11-30, as a reference for randomly selected 20 subjects, yields moderate performance on both Subj 1 and Subj 2.
This demonstrates that the similarity between subjects can largely affect the generalization performance. 
Even with 20 subjects in training, if the subjects are highly dissimilar, the model can achieve little generalization capability, such as 2\% TOP1 Acc. on Subj 2.
We also train models on a mixed set of 20 similar and 20 dissimilar subjects for Subj 1 and Subj 2. 
The results closely match the performance of models trained on the 20 similar subjects alone.

The above experimental results show that when the subjects are similar, the models achieve better generalization performance, and vice versa. 
%
On the other hand, when a mix of similar and dissimilar subjects are used for training, generalization remains stable, with performance approaching to the models trained on similar subjects. 
This suggests that generalization capability depends on learning inherent commonalities among human brains, with substantial tolerance for dissimilarities. 
It also explains why increasing the number of subjects enhances generalization — a larger dataset are more likely to include subjects with higher similarities.

\begin{table}[t]
\setlength\tabcolsep{3pt}
\renewcommand\arraystretch{1.3}
\caption{Results of models trained on subjects that have different gender or similarity. All the training subjects are from Subj 3-167. Best and worst results are marked by {\color{red}red} and {\color{blue}blue}.}
\label{table:similar}
\centering
\scalebox{0.97}{
\begin{tabular}{cccccc}

\hline

Training Subjects &  \makecell{TOP1 Acc. on\\Subj 1 (male)}  &  \makecell{TOP3 Acc. on\\Subj 1 (male)} &   \makecell{TOP1 Acc. on\\Subj 2 (female)} & \makecell{TOP3 Acc. on\\Subj 2 (female)} \\ \hline
50 male & {\color{red}36\%}  & {\color{red}60\%} & {\color{blue}25\%} & {\color{blue}37\%}  \\
50 female & {\color{blue}27\%}  & {\color{blue}40\% } & {\color{red}30\%} & {\color{red}42\%} \\
25 male + 25 female & 32\%  & 49\% & 27\% & 40\% \\
 \hline
20 similar to Subj 1 & {\color{red}21\%}  & {\color{red}36\%} & - & -\\
20 dissimilar to Subj 1 & {\color{blue}8\%}  & {\color{blue}14\%} & - & - \\ \cdashline{1-5}
20 similar to Subj 2 & - & -& {\color{red}21\%}  & {\color{red}31\%}  \\
20 dissimilar to Subj 2 & - & -& {\color{blue}2\%}  & {\color{blue}7\%}  \\\cdashline{1-5}
Subjs 11-30 (20) & 17\% & 31\% & 18\%  & 23\%  \\
 \hline
20 similar + 20 dissimilar to Subj 1 &  21\% & 30\% & - & - \\
20 similar + 20 dissimilar to Subj 2 &  -    & -    & 20\% & 28\%  \\  \hline
\end{tabular}
}
\end{table}

\subsection{Training Strategy}
\label{sec:Training strategy}

\begin{table}[t]
\setlength\tabcolsep{3pt}
\renewcommand\arraystretch{1.3}
\caption{Results of models trained with different training strategies. TOP1 Acc. is averaged over unseen subjects (Subjs 1-10). TOP1 Acc. (seen) is averaged over all seen subjects during training.}
\label{table:str}
\centering
\scalebox{1}{
\begin{tabular}{cccccc}
\hline
Trainging Subjects & Subject Number & Strategies & TOP1 Acc.  & TOP1 Acc. (seen) \\ \hline
Subj 11  & 1 & CLIP / BiMixCo+SoftCLIP & 2\% / 6\%  & 79\% / 83\% \\
Subjs 11-177 & 167 & CLIP / BiMixCo+SoftCLIP & 45\% / 50\%  & 82\% / 84\% \\
 \hline
\end{tabular}
}
\end{table}

In our main experiments, in order to prove that the generalization capability does not come from some specific strategies, we use the simplest contrastive learning pipeline and the CLIP loss to verify our approach on commonly used network architectures.
In this section, we demonstrate that the model performance can be further enhanced if stronger training strategies can be employed. In particular, we adopt the BiMixCo+SoftCLIP training strategy from MindEye1 \citep{scotti2024reconstructing}. 
The BiMixCo+SoftCLIP strategy incorporates a data augmentation technique, extending the mixup approach with the InfoNCE loss \citep{he2020momentum}. 
Additionally, we replace the CLIP loss with the SoftCLIP loss, which leverages softmax probability distributions rather than hard labels. 
Details of these methods can be found in the \textbf{appendix}. 

As shown in Tab. \ref{table:str}, the adoption of BiMixCo+SoftCLIP leads to a noticeable improvement in generalization performance for both single-subject models (TOP1 accuracy improves from 2\% to 6\%) and multiple-subject models (TOP1 accuracy improves from 45\% to 50\%). 
This strategy also enhances the retrieval accuracy on seen subjects. 
These results suggest that better learning strategies can be designed to boost the model generalization capabilities, highlighting the potential of our approach for visual brain decoding.

\subsection{Experimental Results on the NSD Dataset}
\label{sec:Results on NSD dataset}

\begin{table}[t]
\setlength\tabcolsep{3pt}
\renewcommand\arraystretch{1.3}
\caption{Results of models trained on NSD dataset. As in previous works, we randomly extracted 300 pairs from 1,000 pairs of the test set to perform the testing, and run 30 times the experiments to take the average. `MindEye1' means our implemented model of MindEye1 \citep{scotti2024reconstructing}. }
\label{table:nsd}
\centering
\scalebox{0.915}{
\begin{tabular}{ccc|cc|cc}

\hline

&&&\multicolumn{2}{c|}{NSDGeneral Data} & \multicolumn{2}{c}{NSD Whole-brain Data} \\ \hline

Models &Training Subjects & Testing Subjects  & Data Format  & TOP1 Acc.  & Data Format  & TOP1 Acc.  \\ \hline
MindEye1 & Subj 1 & Subj 1 & 1 × 15,724   & 85\%  & 83 × 104 × 81 & 35\%  \\
Ours & Subj 1 & Subj 1 & 1 × 18,000   & 86\%    & 113 × 136 × 113 & 46\%  \\
MindEye1 & Subj 7 & Subj 7  & 1 × 12,682  & 70\%   & 81 × 95 × 78  & 23\%  \\
Ours & Subj 7 & Subj 7  & 1 × 18,000  & 70\%   & 113 × 136 × 113  & 29\%  \\
\hline
Ours & Subjs 1,3,4,6,7,8 & Subj 1 & 1 × 18,000  & 83\%   & 113 × 136 × 113 & 49\% \\
Ours & Subjs 1,3,4,6,7,8 & Subj 7 & 1 × 18,000  & 69\%   & 113 × 136 × 113 & 35\% \\
Ours & Subjs 1,3,4,6,7,8 & Subjs 2,5 & 1 × 18,000  & 1\%   & 113 × 136 × 113 & 1\% \\
 \hline
\end{tabular}
}
\vspace{-0.3cm}
\end{table}

To demonstrate the flexibility of our pipeline, we also train the model on the NSD dataset, including both NSDGeneral and whole-brain data.
The results are shown in Tab. \ref{table:nsd}. We see that the models trained on Subj 1 using NSDGeneral data with both the original data format (\ie, 1 × 15,724 in MindEye1) and our normalized data format (\ie, 1 × 18,000) achieve similar TOP1 accuracy, \ie, 85\% and 86\%, respectively.
(The TOP1 accuracy reported in the original paper of MindEye1 \citep{scotti2024reconstructing} is 84\%.)
The results on Subj 7 can yield similar conclusion.
This demonstrates that our pipeline can be well applied to the individual-specific scenario using NSDGeneral data.
However, by using simple interpolation based upsampling to normalize the data format as a 1 × 18,000 vector, we can train models on multiple subjects with different original NSDGeneral data sizes. As shown in the bottom three rows of Tab. \ref{table:nsd}, by training on subjs 1,3,4,6,7,8 and testing on subj 1 or subj 7, 83\% and 69\% TOP1 accuracy can still be obtained since subj 1 or subj 7 are included in the training data. However, when testing on the unseen subjs 2 and 5, only 1\% TOP1 accuracy is obtained. This is because the NSD dataset has only 8 subjects in total, which is too few to ensure the model generalization performance.  

Let's then evaluate the models trained with whole-brain data. The up right panel of Tab. \ref{table:nsd} shows the results on Subj 1 and Subj 7 by MindEye1, which uses the original whole-brain data format, and our model, which uses the normalized data format (\ie, 113 × 136 × 113). 
We see that in the case of whole-brain data, the simple normalization of data size can improve much the TOP1 accuracy from 35\% to 46\% for subj 1 and from 23\% to 29\% for subj 7. The accuracy is lower than that on NSDGeneral data because the NSDGeneral data are manually labeled brain visual regions.
Again, the normalized data size enables us to train models on multiple subjects with different original data sizes, which cannot be done by MindEye1.
As shown in the bottom right panel of Tab. \ref{table:nsd}, our model trained on six subjects achieves 49\% TOP1 accuracy on Subj 1 and 35\% on Subj 7, outperforming single-subject models trained on Subj 1 (46\%) and Subj 7 (29\%), respectively. 
In contrast, on NSDGeneral data, the multiple-subject models perform slightly worse than their single-subject counterparts.
This discrepancy highlights the individual-specific nature of NSDGeneral data, as discussed in Sec. \ref{sec: dataset} and shown in Fig. \ref{fig:nsdgeneral}. 
We also report the generalization performance of models trained on the six subjects on unseen Subjs 2 and 5. 
Again, the model shows weak generalization ability due to the small number of training subjects in NSD dataset. 

\section{Conclusion}
\label{sec:Conclusion}

Previous visual brain decoding studies typically focused on individual subjects, or training with multiple-subjects but decoding on seen subjects, while little work has been done on exploring the possibility of generalizing visual brain decoding to unseen subjects.
We made an attempt to achieve this goal by leveraging a large dataset from the Human Connectome Project (HCP), constructing 177 × 3,127 image-fMRI pairs from 177 subjects. 
Using this dataset, we proposed a learning paradigm, which utilized whole-brain data and a simple and uniform pipeline for processing all subjects, without requiring individual-specific adaptations.
Via extensive experiments, we found that the model generalization capability emerged with the increase of training subjects, and such generalization capability held across different network architectures. In addition, the similarity between subjects also played a role in improving the generalization capability.
These findings revealed the inherent similarity in brain activities across individuals, which has significant implications for future studies.
As larger, more diverse datasets may become available, this work can provide basis for training a brain encoding foundation model in the future.

\bibliography{iclr2025_tgb}

\begin{thebibliography}{52}
\providecommand{\natexlab}[1]{#1}
\providecommand{\url}[1]{\texttt{#1}}
\expandafter\ifx\csname urlstyle\endcsname\relax
  \providecommand{\doi}[1]{doi: #1}\else
  \providecommand{\doi}{doi: \begingroup \urlstyle{rm}\Url}\fi

\bibitem[Allen et~al.(2022)Allen, St-Yves, Wu, Breedlove, Prince, Dowdle, Nau, Caron, Pestilli, Charest, et~al.]{nsd}
Emily~J Allen, Ghislain St-Yves, Yihan Wu, Jesse~L Breedlove, Jacob~S Prince, Logan~T Dowdle, Matthias Nau, Brad Caron, Franco Pestilli, Ian Charest, et~al.
\newblock A massive 7t fmri dataset to bridge cognitive neuroscience and artificial intelligence.
\newblock \emph{Nature neuroscience}, 25\penalty0 (1):\penalty0 116--126, 2022.

\bibitem[Benchetrit et~al.(2023)Benchetrit, Banville, and King]{benchetrit2023brain}
Yohann Benchetrit, Hubert Banville, and Jean-R{\'e}mi King.
\newblock Brain decoding: toward real-time reconstruction of visual perception.
\newblock \emph{arXiv preprint arXiv:2310.19812}, 2023.

\bibitem[Busch et~al.(2021)Busch, Slipski, Feilong, Guntupalli, di~Oleggio~Castello, Huckins, Nastase, Gobbini, Wager, and Haxby]{busch2021hybrid}
Erica~L Busch, Lukas Slipski, Ma~Feilong, J~Swaroop Guntupalli, Matteo~Visconti di~Oleggio~Castello, Jeremy~F Huckins, Samuel~A Nastase, M~Ida Gobbini, Tor~D Wager, and James~V Haxby.
\newblock Hybrid hyperalignment: A single high-dimensional model of shared information embedded in cortical patterns of response and functional connectivity.
\newblock \emph{NeuroImage}, 233:\penalty0 117975, 2021.

\bibitem[Chang et~al.(2019)Chang, Pyles, Marcus, Gupta, Tarr, and Aminoff]{chang2019bold5000}
Nadine Chang, John~A Pyles, Austin Marcus, Abhinav Gupta, Michael~J Tarr, and Elissa~M Aminoff.
\newblock Bold5000, a public fmri dataset while viewing 5000 visual images.
\newblock \emph{Scientific data}, 6\penalty0 (1):\penalty0 49, 2019.

\bibitem[Chen et~al.(2015)Chen, Chen, Yeshurun, Hasson, Haxby, and Ramadge]{chen2015reduced}
Po-Hsuan~Cameron Chen, Janice Chen, Yaara Yeshurun, Uri Hasson, James Haxby, and Peter~J Ramadge.
\newblock A reduced-dimension fmri shared response model.
\newblock \emph{Advances in neural information processing systems}, 28, 2015.

\bibitem[Chen et~al.(2023)Chen, Qing, Xiang, Yue, and Zhou]{chen2023seeing}
Zijiao Chen, Jiaxin Qing, Tiange Xiang, Wan~Lin Yue, and Juan~Helen Zhou.
\newblock Seeing beyond the brain: Conditional diffusion model with sparse masked modeling for vision decoding.
\newblock In \emph{Proceedings of the IEEE/CVF Conference on Computer Vision and Pattern Recognition}, pp.\  22710--22720, 2023.

\bibitem[Du et~al.(2023)Du, Fu, Li, and He]{du2023decoding}
Changde Du, Kaicheng Fu, Jinpeng Li, and Huiguang He.
\newblock Decoding visual neural representations by multimodal learning of brain-visual-linguistic features.
\newblock \emph{IEEE Transactions on Pattern Analysis and Machine Intelligence}, 45\penalty0 (9):\penalty0 10760--10777, 2023.

\bibitem[Ferrante et~al.(2024)Ferrante, Boccato, Ozcelik, VanRullen, and Toschi]{Throughtheireyes}
Matteo Ferrante, Tommaso Boccato, Furkan Ozcelik, Rufin VanRullen, and Nicola Toschi.
\newblock Through their eyes: multi-subject brain decoding with simple alignment techniques.
\newblock \emph{Imaging Neuroscience}, 2:\penalty0 1--21, 2024.

\bibitem[Gu et~al.(2022)Gu, Jamison, Kuceyeski, and Sabuncu]{gu2022decoding}
Zijin Gu, Keith Jamison, Amy Kuceyeski, and Mert Sabuncu.
\newblock Decoding natural image stimuli from fmri data with a surface-based convolutional network.
\newblock \emph{arXiv preprint arXiv:2212.02409}, 2022.

\bibitem[Han et~al.(2024)Han, Lee, and Ye]{han2024mindformer}
Inhwa Han, Jaayeon Lee, and Jong~Chul Ye.
\newblock Mindformer: A transformer architecture for multi-subject brain decoding via fmri.
\newblock \emph{arXiv preprint arXiv:2405.17720}, 2024.

\bibitem[Hanke et~al.(2014)Hanke, Baumgartner, Ibe, Kaule, Pollmann, Speck, Zinke, and Stadler]{hanke2014high}
Michael Hanke, Florian~J Baumgartner, Pierre Ibe, Falko~R Kaule, Stefan Pollmann, Oliver Speck, Wolf Zinke, and J{\"o}rg Stadler.
\newblock A high-resolution 7-tesla fmri dataset from complex natural stimulation with an audio movie.
\newblock \emph{Scientific data}, 1\penalty0 (1):\penalty0 1--18, 2014.

\bibitem[Haxby et~al.(2011)Haxby, Guntupalli, Connolly, Halchenko, Conroy, Gobbini, Hanke, and Ramadge]{haxby2011common}
James~V Haxby, J~Swaroop Guntupalli, Andrew~C Connolly, Yaroslav~O Halchenko, Bryan~R Conroy, M~Ida Gobbini, Michael Hanke, and Peter~J Ramadge.
\newblock A common, high-dimensional model of the representational space in human ventral temporal cortex.
\newblock \emph{Neuron}, 72\penalty0 (2):\penalty0 404--416, 2011.

\bibitem[Haxby et~al.(2020)Haxby, Guntupalli, Nastase, and Feilong]{haxby2020hyperalignment}
James~V Haxby, J~Swaroop Guntupalli, Samuel~A Nastase, and Ma~Feilong.
\newblock Hyperalignment: Modeling shared information encoded in idiosyncratic cortical topographies.
\newblock \emph{elife}, 9:\penalty0 e56601, 2020.

\bibitem[He et~al.(2016)He, Zhang, Ren, and Sun]{ResNet}
Kaiming He, Xiangyu Zhang, Shaoqing Ren, and Jian Sun.
\newblock Deep residual learning for image recognition.
\newblock In \emph{Proceedings of the IEEE Conference on Computer Vision and Pattern Recognition (CVPR)}, June 2016.

\bibitem[He et~al.(2020)He, Fan, Wu, Xie, and Girshick]{he2020momentum}
Kaiming He, Haoqi Fan, Yuxin Wu, Saining Xie, and Ross Girshick.
\newblock Momentum contrast for unsupervised visual representation learning.
\newblock In \emph{Proceedings of the IEEE/CVF conference on computer vision and pattern recognition}, pp.\  9729--9738, 2020.

\bibitem[Ho et~al.(2020)Ho, Jain, and Abbeel]{ho2020denoising}
Jonathan Ho, Ajay Jain, and Pieter Abbeel.
\newblock Denoising diffusion probabilistic models.
\newblock \emph{Advances in neural information processing systems}, 33:\penalty0 6840--6851, 2020.

\bibitem[Horikawa \& Kamitani(2017)Horikawa and Kamitani]{horikawa2017generic}
Tomoyasu Horikawa and Yukiyasu Kamitani.
\newblock Generic decoding of seen and imagined objects using hierarchical visual features.
\newblock \emph{Nature communications}, 8\penalty0 (1):\penalty0 15037, 2017.

\bibitem[Huang et~al.(2022)Huang, Busch, Wallenstein, Gerasimiuk, Benz, Lajoie, Wolf, Turk-Browne, and Krishnaswamy]{huang2022learning}
Jessie Huang, Erica Busch, Tom Wallenstein, Michal Gerasimiuk, Andrew Benz, Guillaume Lajoie, Guy Wolf, Nicholas Turk-Browne, and Smita Krishnaswamy.
\newblock Learning shared neural manifolds from multi-subject fmri data.
\newblock In \emph{2022 IEEE 32nd International Workshop on Machine Learning for Signal Processing (MLSP)}, pp.\  01--06. IEEE, 2022.

\bibitem[Jenkinson et~al.(2002)Jenkinson, Bannister, Brady, and Smith]{jenkinson2002improved}
Mark Jenkinson, Peter Bannister, Michael Brady, and Stephen Smith.
\newblock Improved optimization for the robust and accurate linear registration and motion correction of brain images.
\newblock \emph{Neuroimage}, 17\penalty0 (2):\penalty0 825--841, 2002.

\bibitem[Jiang et~al.(2024)Jiang, Meng, Liu, Li, Su, and Zhao]{jiang2024mindshot}
Shuai Jiang, Zhu Meng, Delong Liu, Haiwen Li, Fei Su, and Zhicheng Zhao.
\newblock Mindshot: Brain decoding framework using only one image.
\newblock \emph{arXiv preprint arXiv:2405.15278}, 2024.

\bibitem[Kamitani \& Tong(2005)Kamitani and Tong]{kamitani2005decoding}
Yukiyasu Kamitani and Frank Tong.
\newblock Decoding the visual and subjective contents of the human brain.
\newblock \emph{Nature neuroscience}, 8\penalty0 (5):\penalty0 679--685, 2005.

\bibitem[Kaur \& Gandhi(2019)Kaur and Gandhi]{kaur2019automated}
Taranjit Kaur and Tapan~Kumar Gandhi.
\newblock Automated brain image classification based on vgg-16 and transfer learning.
\newblock In \emph{2019 international conference on information technology (ICIT)}, pp.\  94--98. IEEE, 2019.

\bibitem[Kay et~al.(2008)Kay, Naselaris, Prenger, and Gallant]{kay2008identifying}
Kendrick~N Kay, Thomas Naselaris, Ryan~J Prenger, and Jack~L Gallant.
\newblock Identifying natural images from human brain activity.
\newblock \emph{Nature}, 452\penalty0 (7185):\penalty0 352--355, 2008.

\bibitem[Khosla et~al.(2020)Khosla, Ngo, Jamison, Kuceyeski, and Sabuncu]{khosla2020}
Meenakshi Khosla, Gia~H Ngo, Keith Jamison, Amy Kuceyeski, and Mert~R Sabuncu.
\newblock A shared neural encoding model for the prediction of subject-specific fmri response.
\newblock In \emph{Medical Image Computing and Computer Assisted Intervention--MICCAI 2020: 23rd International Conference, Lima, Peru, October 4--8, 2020, Proceedings, Part VII 23}, pp.\  539--548. Springer, 2020.

\bibitem[Kim et~al.(2020)Kim, Lee, Bae, and Yun]{Kim2020MixCoMC}
Sungnyun Kim, Gihun Lee, Sangmin Bae, and Seyoung Yun.
\newblock Mixco: Mix-up contrastive learning for visual representation.
\newblock \emph{ArXiv}, abs/2010.06300, 2020.

\bibitem[Lorbert \& Ramadge(2012)Lorbert and Ramadge]{lorbert2012kernel}
Alexander Lorbert and Peter~J Ramadge.
\newblock Kernel hyperalignment.
\newblock \emph{Advances in Neural Information Processing Systems}, 25, 2012.

\bibitem[Loshchilov \& Hutter(2017)Loshchilov and Hutter]{adamw}
Ilya Loshchilov and Frank Hutter.
\newblock Decoupled weight decay regularization.
\newblock \emph{arXiv preprint arXiv:1711.05101}, 2017.

\bibitem[Lu et~al.(2024)Lu, Du, Wang, Zhu, Jiang, and He]{lu2024animate}
Yizhuo Lu, Changde Du, Chong Wang, Xuanliu Zhu, Liuyun Jiang, and Huiguang He.
\newblock Animate your thoughts: Decoupled reconstruction of dynamic natural vision from slow brain activity.
\newblock \emph{arXiv preprint arXiv:2405.03280}, 2024.

\bibitem[Matteo et~al.(2020)Matteo, Chauhan, Jiahui, and Gobbini]{visconti2020fmri}
Visconti di Oleggio~Castello Matteo, Vassiki Chauhan, Guo Jiahui, and M~Ida Gobbini.
\newblock An fmri dataset in response to “the grand budapest hotel”, a socially-rich, naturalistic movie.
\newblock \emph{Scientific Data}, 7\penalty0 (1):\penalty0 383, 2020.

\bibitem[Matteo et~al.(2021)Matteo, Haxby, and Gobbini]{visconti2021shared}
Visconti di Oleggio~Castello Matteo, James~V Haxby, and M~Ida Gobbini.
\newblock Shared neural codes for visual and semantic information about familiar faces in a common representational space.
\newblock \emph{Proceedings of the National Academy of Sciences}, 118\penalty0 (45):\penalty0 e2110474118, 2021.

\bibitem[Naselaris et~al.(2011)Naselaris, Kay, Nishimoto, and Gallant]{naselaris2011encoding}
Thomas Naselaris, Kendrick~N Kay, Shinji Nishimoto, and Jack~L Gallant.
\newblock Encoding and decoding in fmri.
\newblock \emph{Neuroimage}, 56\penalty0 (2):\penalty0 400--410, 2011.

\bibitem[Ozcelik \& VanRullen(2023)Ozcelik and VanRullen]{braindiffuser}
Furkan Ozcelik and Rufin VanRullen.
\newblock Natural scene reconstruction from fmri signals using generative latent diffusion.
\newblock \emph{Scientific Reports}, 13\penalty0 (1):\penalty0 15666, 2023.

\bibitem[Paszke et~al.(2017)Paszke, Gross, Chintala, Chanan, Yang, DeVito, Lin, Desmaison, Antiga, and Lerer]{pytorch}
Adam Paszke, Sam Gross, Soumith Chintala, Gregory Chanan, Edward Yang, Zachary DeVito, Zeming Lin, Alban Desmaison, Luca Antiga, and Adam Lerer.
\newblock Automatic differentiation in pytorch.
\newblock 2017.

\bibitem[Prince et~al.(2022)Prince, Charest, Kurzawski, Pyles, Tarr, and Kay]{prince2022improving}
Jacob~S Prince, Ian Charest, Jan~W Kurzawski, John~A Pyles, Michael~J Tarr, and Kendrick~N Kay.
\newblock Improving the accuracy of single-trial fmri response estimates using glmsingle.
\newblock \emph{Elife}, 11:\penalty0 e77599, 2022.

\bibitem[Qian et~al.(2023)Qian, Wang, Huo, Feng, and Fu]{qian2023fmri}
Xuelin Qian, Yun Wang, Jingyang Huo, Jianfeng Feng, and Yanwei Fu.
\newblock fmri-pte: A large-scale fmri pretrained transformer encoder for multi-subject brain activity decoding.
\newblock \emph{arXiv preprint arXiv:2311.00342}, 2023.

\bibitem[{Radford} et~al.(2021){Radford}, {Kim}, {Hallacy}, {Ramesh}, {Goh}, {Agarwal}, {Sastry}, {Askell}, {Mishkin}, {Clark}, {Krueger}, and {Sutskever}]{CLIP}
Alec {Radford}, Jong~Wook {Kim}, Chris {Hallacy}, Aditya {Ramesh}, Gabriel {Goh}, Sandhini {Agarwal}, Girish {Sastry}, Amanda {Askell}, Pamela {Mishkin}, Jack {Clark}, Gretchen {Krueger}, and Ilya {Sutskever}.
\newblock {Learning Transferable Visual Models From Natural Language Supervision}.
\newblock \emph{arXiv e-prints}, art. arXiv:2103.00020, February 2021.

\bibitem[Rombach et~al.(2022)Rombach, Blattmann, Lorenz, Esser, and Ommer]{SD}
Robin Rombach, Andreas Blattmann, Dominik Lorenz, Patrick Esser, and Bj{\"o}rn Ommer.
\newblock High-resolution image synthesis with latent diffusion models.
\newblock In \emph{Proceedings of the IEEE/CVF conference on computer vision and pattern recognition}, pp.\  10684--10695, 2022.

\bibitem[Schirrmeister et~al.(2017)Schirrmeister, Springenberg, Fiederer, Glasstetter, Eggensperger, Tangermann, Hutter, Burgard, and Ball]{schirrmeister2017deep}
Robin~Tibor Schirrmeister, Jost~Tobias Springenberg, Lukas Dominique~Josef Fiederer, Martin Glasstetter, Katharina Eggensperger, Michael Tangermann, Frank Hutter, Wolfram Burgard, and Tonio Ball.
\newblock Deep learning with convolutional neural networks for eeg decoding and visualization.
\newblock \emph{Human brain mapping}, 38\penalty0 (11):\penalty0 5391--5420, 2017.

\bibitem[Schneider et~al.(2023)Schneider, Lee, and Mathis]{schneider2023learnable}
Steffen Schneider, Jin~Hwa Lee, and Mackenzie~Weygandt Mathis.
\newblock Learnable latent embeddings for joint behavioural and neural analysis.
\newblock \emph{Nature}, 617\penalty0 (7960):\penalty0 360--368, 2023.

\bibitem[Scotti et~al.(2024{\natexlab{a}})Scotti, Banerjee, Goode, Shabalin, Nguyen, Dempster, Verlinde, Yundler, Weisberg, Norman, et~al.]{scotti2024reconstructing}
Paul Scotti, Atmadeep Banerjee, Jimmie Goode, Stepan Shabalin, Alex Nguyen, Aidan Dempster, Nathalie Verlinde, Elad Yundler, David Weisberg, Kenneth Norman, et~al.
\newblock Reconstructing the mind's eye: fmri-to-image with contrastive learning and diffusion priors.
\newblock \emph{Advances in Neural Information Processing Systems}, 36, 2024{\natexlab{a}}.

\bibitem[Scotti et~al.(2024{\natexlab{b}})Scotti, Tripathy, Villanueva, Kneeland, Chen, Narang, Santhirasegaran, Xu, Naselaris, Norman, et~al.]{scotti2024mindeye2}
Paul~S Scotti, Mihir Tripathy, Cesar Kadir~Torrico Villanueva, Reese Kneeland, Tong Chen, Ashutosh Narang, Charan Santhirasegaran, Jonathan Xu, Thomas Naselaris, Kenneth~A Norman, et~al.
\newblock Mindeye2: Shared-subject models enable fmri-to-image with 1 hour of data.
\newblock \emph{arXiv preprint arXiv:2403.11207}, 2024{\natexlab{b}}.

\bibitem[Sexton \& Love(2022)Sexton and Love]{sexton2022reassessing}
Nicholas~J Sexton and Bradley~C Love.
\newblock Reassessing hierarchical correspondences between brain and deep networks through direct interface.
\newblock \emph{Science advances}, 8\penalty0 (28):\penalty0 eabm2219, 2022.

\bibitem[Shvartsman et~al.(2018)Shvartsman, Sundaram, Aoi, Charles, Willke, and Cohen]{shvartsman2018matrix}
Michael Shvartsman, Narayanan Sundaram, Mikio Aoi, Adam Charles, Theodore Willke, and Jonathan Cohen.
\newblock Matrix-normal models for fmri analysis.
\newblock In \emph{International conference on artificial intelligence and statistics}, pp.\  1914--1923. PMLR, 2018.

\bibitem[Sohl-Dickstein et~al.(2015)Sohl-Dickstein, Weiss, Maheswaranathan, and Ganguli]{sohl2015deep}
Jascha Sohl-Dickstein, Eric Weiss, Niru Maheswaranathan, and Surya Ganguli.
\newblock Deep unsupervised learning using nonequilibrium thermodynamics.
\newblock In \emph{International conference on machine learning}, pp.\  2256--2265. PMLR, 2015.

\bibitem[Takagi \& Nishimoto(2023)Takagi and Nishimoto]{takagi2023high}
Yu~Takagi and Shinji Nishimoto.
\newblock High-resolution image reconstruction with latent diffusion models from human brain activity.
\newblock In \emph{Proceedings of the IEEE/CVF Conference on Computer Vision and Pattern Recognition}, pp.\  14453--14463, 2023.

\bibitem[Vallabhaneni et~al.(2021)Vallabhaneni, Sharma, Kumar, Kulshreshtha, Reddy, Kumar, Kumar, and Bitra]{vallabhaneni2021deep}
Ramesh~Babu Vallabhaneni, Pankaj Sharma, Vinit Kumar, Vyom Kulshreshtha, Koya~Jeevan Reddy, S~Selva Kumar, V~Sandeep Kumar, and Surendra~Kumar Bitra.
\newblock Deep learning algorithms in eeg signal decoding application: a review.
\newblock \emph{IEEE Access}, 9:\penalty0 125778--125786, 2021.

\bibitem[Van~Essen et~al.(2013)Van~Essen, Smith, Barch, Behrens, Yacoub, Ugurbil, Consortium, et~al.]{HCP}
David~C Van~Essen, Stephen~M Smith, Deanna~M Barch, Timothy~EJ Behrens, Essa Yacoub, Kamil Ugurbil, Wu-Minn~HCP Consortium, et~al.
\newblock The wu-minn human connectome project: an overview.
\newblock \emph{Neuroimage}, 80:\penalty0 62--79, 2013.

\bibitem[Vaswani et~al.(2017)Vaswani, Shazeer, Parmar, Uszkoreit, Jones, Gomez, Kaiser, and Polosukhin]{Transformer}
Ashish Vaswani, Noam Shazeer, Niki Parmar, Jakob Uszkoreit, Llion Jones, Aidan~N Gomez, {\L}ukasz Kaiser, and Illia Polosukhin.
\newblock Attention is all you need.
\newblock \emph{Advances in neural information processing systems}, 30, 2017.

\bibitem[Wagner et~al.(2022)Wagner, Waite, Wierzba, Hoffstaedter, Waite, Poldrack, Eickhoff, and Hanke]{wagner2022fairly}
Adina~S Wagner, Laura~K Waite, Ma{\l}gorzata Wierzba, Felix Hoffstaedter, Alexander~Q Waite, Benjamin Poldrack, Simon~B Eickhoff, and Michael Hanke.
\newblock Fairly big: A framework for computationally reproducible processing of large-scale data.
\newblock \emph{Scientific data}, 9\penalty0 (1):\penalty0 80, 2022.

\bibitem[Xia et~al.(2024)Xia, de~Charette, {\"O}ztireli, and Xue]{umbrae}
Weihao Xia, Raoul de~Charette, Cengiz {\"O}ztireli, and Jing-Hao Xue.
\newblock Umbrae: Unified multimodal decoding of brain signals.
\newblock \emph{arXiv preprint arXiv:2404.07202}, 2024.

\bibitem[Xu et~al.(2012)Xu, Lorbert, Ramadge, Guntupalli, and Haxby]{xu2012regularized}
Hao Xu, Alexander Lorbert, Peter~J Ramadge, J~Swaroop Guntupalli, and James~V Haxby.
\newblock Regularized hyperalignment of multi-set fmri data.
\newblock In \emph{2012 IEEE statistical signal processing workshop (SSP)}, pp.\  229--232. IEEE, 2012.

\bibitem[Zhou et~al.(2024)Zhou, Du, Wang, and He]{CLIP-MUSED}
Qiongyi Zhou, Changde Du, Shengpei Wang, and Huiguang He.
\newblock Clip-mused: Clip-guided multi-subject visual neural information semantic decoding.
\newblock \emph{arXiv preprint arXiv:2402.08994}, 2024.

\end{thebibliography}
\bibliographystyle{iclr2025_conference}

\clearpage

\renewcommand\thesection{\Alph{section}}
	\renewcommand\thesubsection{\thesection.\arabic{subsection}}
	\renewcommand\thefigure{\Alph{section}.\arabic{figure}}
	\renewcommand\thetable{\Alph{section}.\arabic{table}} 
	
	\noindent{\LARGE{\textbf{Appendix}}}
	
	\setcounter{section}{0}
	\setcounter{figure}{0}
	\setcounter{table}{0}
\vspace{0.6cm}

In this appendix, we first provide the details of the networks used in the main paper. 
Second, we outline the details of the BiMixCo+SoftCLIP training strategy used in Sec. 4.5 of the main paper.
Lastly, we present the visualization of the image retrieval results.

\section{The Details of Architectures}
\label{sec:Architectures}

In the main paper, we validated the generalization capability of visual brain decoding models on unseen subjects across popular network architectures, including MLP, CNN and Transformer.
We show the detailed network structures in this section.
The input to the networks is fMRI voxels of size 113 × 136 × 113.

\begin{table}[h!]
    \centering
    \caption{Detailed architecture of our MLP network.}
    \label{mlp}
    \scalebox{0.78}
    {
\begin{tabular}{>{\centering\arraybackslash}m{3.5cm} >{\centering\arraybackslash}m{4.5cm} >{\centering\arraybackslash}m{3cm} >{\centering\arraybackslash}m{2.5cm} >{\centering\arraybackslash}m{2cm} }
        \hline
        \textbf{Layer Type} & \textbf{Input Size} & \textbf{Output Size} & \textbf{Kernel/Linear Size} & \textbf{Stride/Padding} \\
        \hline
        Conv3D (conv1)       & $1 \times 113 \times 136 \times 113$ & $32 \times 38 \times 46 \times 38$ & $9 \times 9 \times 9$ & Stride 3, Padding 4 \\
        BatchNorm3D (bn1)    & $32 \times 38 \times 46 \times 38$ & $32 \times 38 \times 46 \times 38$ & - & - \\
        Conv3D (conv2)       & $32 \times 38 \times 46 \times 38$ & $48 \times 19 \times 23 \times 19$ & $7 \times 7 \times 7$ & Stride 2, Padding 3 \\
        BatchNorm3D (bn2)    & $48 \times 19 \times 23 \times 19$ & $48 \times 19 \times 23 \times 19$ & - & - \\
        Conv3D (conv3)       & $48 \times 19 \times 23 \times 19$ & $64 \times 10 \times 12 \times 10$ & $5 \times 5 \times 5$ & Stride 2, Padding 2 \\
        BatchNorm3D (bn3)    & $64 \times 10 \times 12 \times 10$ & $64 \times 10 \times 12 \times 10$ & - & - \\
        ReLU (activation)    & $64 \times 10 \times 12 \times 10$ & $64 \times 10 \times 12 \times 10$ & - & - \\
        Flatten              & $64 \times 10 \times 12 \times 10$ & $76800$ & - & - \\
        Linear (lin0)        & $76800$ & $4096$ & - & - \\
        MLP (4 layers)       & $4096$ & $4096$ & - & - \\
        Linear (lin1)        & $4096$ & $257 \times 1024$ & - & - \\
        \hline
    \end{tabular}
    }
\end{table}

\paragraph{MLP.} 

Tab. \ref{mlp} presents the architecture of our MLP network. 
We first employ a 3-layer 3D CNN for feature extraction, followed by a fully connected layer to reduces the feature to a 1D vector of 4096 elements.
Similar to MindEye1 \citep{scotti2024reconstructing}, we apply the MLP backbone in this feature space with the vector length of 4096.
Finally, we utilize a fully connected layer to expand the feature from 4096 to 257 × 1024, which is used for loss calculation.

\begin{table}[h!]
    \centering
    \caption{Detailed architecture of our 1D CNN network.}
        \label{1d}
        \scalebox{0.78}
    {
    \begin{tabular}{>{\centering\arraybackslash}m{3.5cm} >{\centering\arraybackslash}m{4.5cm} >{\centering\arraybackslash}m{3cm} >{\centering\arraybackslash}m{2.5cm} >{\centering\arraybackslash}m{2cm} }
        \hline
        \textbf{Layer Type} & \textbf{Input Size} & \textbf{Output Size} & \textbf{Kernel/Linear Size} & \textbf{Stride/Padding} \\
        \hline
        Conv3D (conv1)       & $1 \times 113 \times 136 \times 113$ & $32 \times 38 \times 46 \times 38$ & $9 \times 9 \times 9$ & Stride 3, Padding 4 \\
        BatchNorm3D (bn1)    & $32 \times 38 \times 46 \times 38$ & $32 \times 38 \times 46 \times 38$ & - & - \\
        Conv3D (conv2)       & $32 \times 38 \times 46 \times 38$ & $48 \times 19 \times 23 \times 19$ & $7 \times 7 \times 7$ & Stride 2, Padding 3 \\
        BatchNorm3D (bn2)    & $48 \times 19 \times 23 \times 19$ & $48 \times 19 \times 23 \times 19$ & - & - \\
        Conv3D (conv3)       & $48 \times 19 \times 23 \times 19$ & $64 \times 10 \times 12 \times 10$ & $5 \times 5 \times 5$ & Stride 2, Padding 2 \\
        BatchNorm3D (bn3)    & $64 \times 10 \times 12 \times 10$ & $64 \times 10 \times 12 \times 10$ & - & - \\
        ReLU (activation)    & $64 \times 10 \times 12 \times 10$ & $64 \times 10 \times 12 \times 10$ & - & - \\
        Flatten              & $64 \times 10 \times 12 \times 10$ & $76800$ & - & - \\
        Linear (lin0)        & $76800$ & $4096$ & - & - \\
        Reshape              & $4096$ & $1 \times 4096$ & - & - \\
        Conv1D (conv01)      & $1 \times 4096$ & $64 \times 680$ & Kernel 13 & Stride 6 \\
        BatchNorm1D (bn01)   & $64 \times 680$ & $64 \times 680$ & - & - \\
        Conv1D (conv02)      & $64 \times 680$ & $128 \times 134$ & Kernel 11 & Stride 5 \\
        BatchNorm1D (bn02)   & $128 \times 134$ & $128 \times 134$ & - & - \\
        Conv1D (conv03)      & $128 \times 134$ & $256 \times 32$ & Kernel 9 & Stride 4 \\
        BatchNorm1D (bn03)   & $256 \times 32$ & $256 \times 32$ & - & - \\
        Conv1D (conv04)      & $256 \times 32$ & $512 \times 9$ & Kernel 7 & Stride 3 \\
        BatchNorm1D (bn04)   & $512 \times 9$ & $512 \times 9$ & - & - \\
        ResBlock1D (x8)      & $512 \times 9$ & $512 \times 9$ & - & - \\
        Flatten              & $512 \times 9$ & $4608$ & - & - \\
        Linear (lin1)        & $4608$ & $ 257 \times 1024$ & - & - \\
        \hline
    \end{tabular}
    }
\end{table}

\paragraph{1D CNN.}

As shown in Tab. \ref{1d}, similar to our MLP network, our 1D CNN network first reduces the feature to a vector of length 4096.
Then we use several separate 1D CNN layers and eight 1D ResBlock layers for feature mapping, which constitute our 1D CNN backbone.
Finally, a fully connected layer is used to produce the output with size of 257 × 1024.

\begin{table}[h!]
    \centering
    \caption{Detailed architecture of our 3D CNN network.}
        \label{3d}
        \scalebox{0.78}
    {
    \begin{tabular}{>{\centering\arraybackslash}m{3.5cm} >{\centering\arraybackslash}m{4.5cm} >{\centering\arraybackslash}m{3cm} >{\centering\arraybackslash}m{2.5cm} >{\centering\arraybackslash}m{2cm} }
        \hline
        \textbf{Layer Type} & \textbf{Input Size} & \textbf{Output Size} & \textbf{Kernel/Linear Size} & \textbf{Stride/Padding} \\
        \hline
        Conv3D (conv1)       & $1 \times 113 \times 136 \times 113$ & $32 \times 38 \times 46 \times 38$ & $9 \times 9 \times 9$ & Stride 3, Padding 4 \\
        BatchNorm3D (bn1)    & $32 \times 38 \times 46 \times 38$ & $32 \times 38 \times 46 \times 38$ & - & - \\
        Conv3D (conv2)       & $32 \times 38 \times 46 \times 38$ & $48 \times 19 \times 23 \times 19$ & $7 \times 7 \times 7$ & Stride 2, Padding 3 \\
        BatchNorm3D (bn2)    & $48 \times 19 \times 23 \times 19$ & $48 \times 19 \times 23 \times 19$ & - & - \\
        Conv3D (conv3)       & $48 \times 19 \times 23 \times 19$ & $64 \times 10 \times 12 \times 10$ & $5 \times 5 \times 5$ & Stride 2, Padding 2 \\
        BatchNorm3D (bn3)    & $64 \times 10 \times 12 \times 10$ & $64 \times 10 \times 12 \times 10$ & - & - \\
        Conv3D (conv4)       & $64 \times 10 \times 12 \times 10$ & $90 \times 5 \times 6 \times 5$ & $5 \times 5 \times 5$ & Stride 2, Padding 2 \\
        BatchNorm3D (bn4)    & $90 \times 5 \times 6 \times 5$ & $90 \times 5 \times 6 \times 5$ & - & - \\
        Conv3D (conv5)       & $90 \times 5 \times 6 \times 5$ & $150 \times 3 \times 3 \times 3$ & $5 \times 5 \times 5$ & Stride 2, Padding 2 \\
        BatchNorm3D (bn5)    & $150 \times 3 \times 3 \times 3$ & $150 \times 3 \times 3 \times 3$ & - & - \\
        ReLU (activation)    & $150 \times 3 \times 3 \times 3$ & $150 \times 3 \times 3 \times 3$ & - & - \\
        ResBlock3D (x8)      & $150 \times 3 \times 3 \times 3$ & $150 \times 3 \times 3 \times 3$ & - & - \\
        Flatten              & $150 \times 3 \times 3 \times 3$ & $4050$ & - & - \\
        Linear (lin1)        & $4050$ & $ 257 \times 1024$ & - & - \\
        \hline
    \end{tabular}
    }
\end{table}

\paragraph{3D CNN.}

Different from the above MLP and 1D CNN networks, as shown in Tab. \ref{3d}, our 3D CNN network uses 3D CNN layers and 3D ResBlock layers to process the input.
After the main processing, a fully connected layer is used to produce the final output with size of 257 × 1024.

\begin{table}[h!]
    \centering
    \caption{Detailed architecture of our Transformer network.}
        \label{tr}
        \scalebox{0.78}
    {
    \begin{tabular}{>{\centering\arraybackslash}m{4cm} >{\centering\arraybackslash}m{4.5cm} >{\centering\arraybackslash}m{3cm} >{\centering\arraybackslash}m{2.5cm} >{\centering\arraybackslash}m{2cm} }
        \hline
        \textbf{Layer Type} & \textbf{Input Size} & \textbf{Output Size} & \textbf{Kernel/Linear Size} & \textbf{Stride/Padding} \\
        \hline
        Conv3D (conv1)       & $1 \times 113 \times 136 \times 113$ & $32 \times 38 \times 46 \times 38$ & $9 \times 9 \times 9$ & Stride 3, Padding 4 \\
        BatchNorm3D (bn1)    & $32 \times 38 \times 46 \times 38$ & $32 \times 38 \times 46 \times 38$ & - & - \\
        Conv3D (conv2)       & $32 \times 38 \times 46 \times 38$ & $48 \times 19 \times 23 \times 19$ & $7 \times 7 \times 7$ & Stride 2, Padding 3 \\
        BatchNorm3D (bn2)    & $48 \times 19 \times 23 \times 19$ & $48 \times 19 \times 23 \times 19$ & - & - \\
        Conv3D (conv3)       & $48 \times 19 \times 23 \times 19$ & $64 \times 10 \times 12 \times 10$ & $5 \times 5 \times 5$ & Stride 2, Padding 2 \\
        BatchNorm3D (bn3)    & $64 \times 10 \times 12 \times 10$ & $64 \times 10 \times 12 \times 10$ & - & - \\
        ReLU (activation)    & $64 \times 10 \times 12 \times 10$ & $64 \times 10 \times 12 \times 10$ & - & - \\
        Flatten              & $64 \times 10 \times 12 \times 10$ & $76800$ & - & - \\
        Linear (lin0)        & $76800$ & $4096$ & - & - \\
        Reshape              & $4096$ & $16 \times 256$ & - & - \\
        Transformer (d=256, h=8) (x24)       & $16 \times 256$ & $16 \times 256$ & - & - \\
        Reshape              & $16 \times 256$ & $4096$ & - & - \\
        Linear (lin1)        & $4096$ & $ 257 \times 1024$ & - & - \\
        
        \hline
    \end{tabular}
    }
\end{table}

\paragraph{Transformer.}

The details of our Transformer network are shown in Tab. \ref{tr}. Similar to our MLP and 1D CNN networks, our Transformer network first reduces the feature to a 4096 sized vector.
Then we reshape the feature to 16 × 256, which is then sent to the Transformer backbone with 24 Transformer block layers.
After that, we reshape the output of Transformer backbone back to 4096 and employ a fully connected layer to produce the final output with size of 257 × 1024.

\begin{figure}[h!]
\begin{center}
\includegraphics[width=0.94\linewidth]{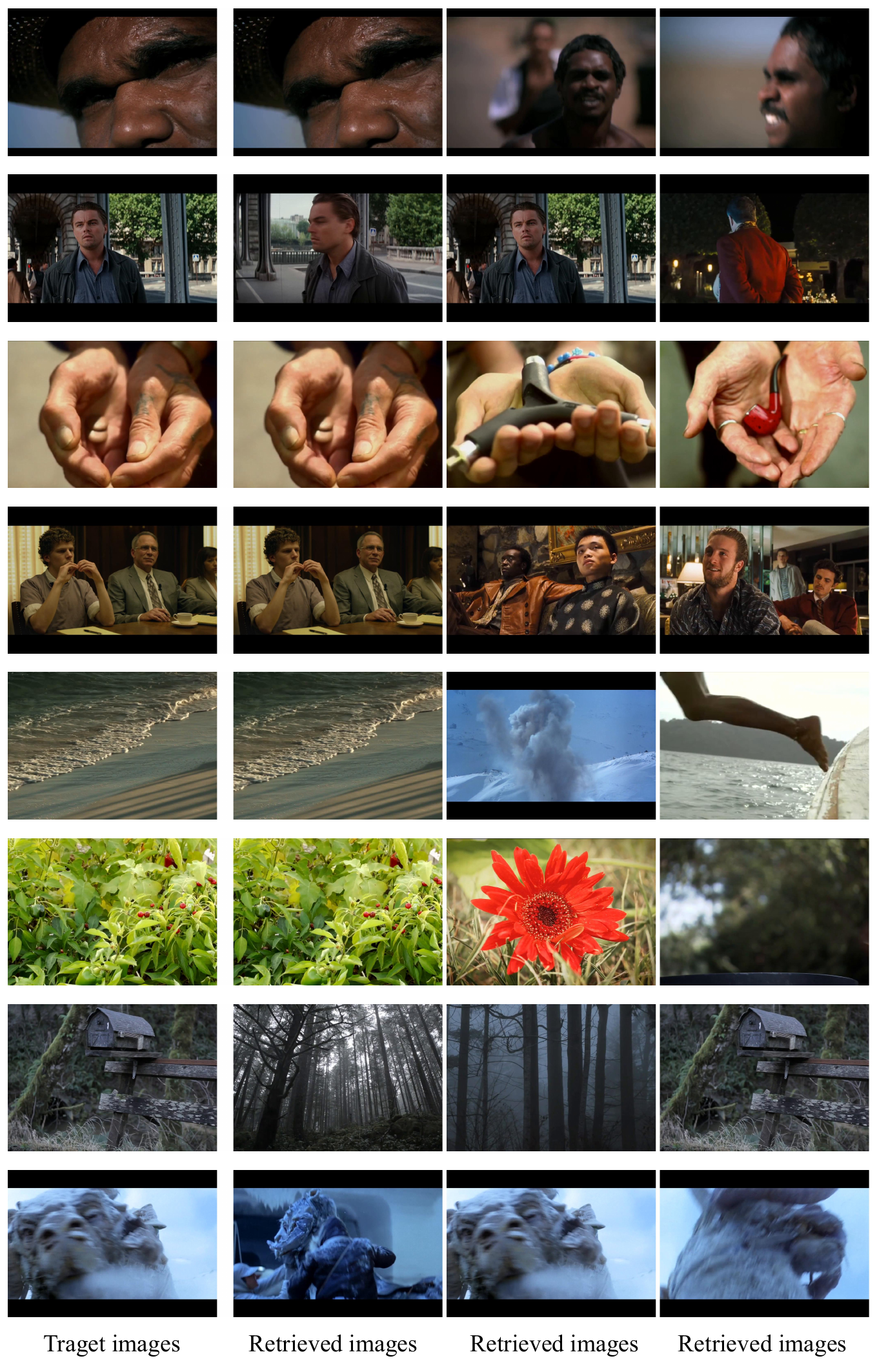}
\end{center}
\vspace{-0.5cm}
\caption{The visualized image retrieval results. `Target images' refer to those viewed by Subj 1 from the HCP dataset, while `Retrieved images' represent the corresponding retrieval outputs from the visual brain decoding model trained on 167 subjects. The retrieved images are ranked from left to right based on retrieval similarity.}
\label{fig:vis}
\end{figure}

\section{The Details of Training Strategy}
\label{sec:Training strategies}

In our main paper, we adopt the BiMixCo+SoftCLIP training strategy from MindEye1 \citep{scotti2024reconstructing} to further enhance our visual brain decoding network. 
In this section, we provide the details of this strategy, which uses BiMixCo for one-third of the training and employs SoftCLIP for the rest of training.
\paragraph{BiMixCo.}
The BiMixCo combines MixCo \citep{Kim2020MixCoMC} (an extension of mixup that uses the InfoNCE loss) and the bidirectional CLIP loss.
To be specific, as described in MindEye1 \citep{scotti2024reconstructing}, voxels are mixed by a factor $\lambda$ sampled from the Beta distribution with $\alpha=\beta=0.15$:
\begin{align}
    x_{\text{mix}_{i,k_i}} = \lambda_i\cdot x_i + (1-\lambda_i)\cdot x_{k_i}, \quad p_i^* = f(x_{\text{mix}_{i,k_i}}), \quad p_i = f(x_i), \quad t_i = \text{CLIP}_\text{Image}(y_i),
\end{align}
where $x_i$ and $y_i$ represent the $i$-th fMRI voxel and image, respectively. $k_i \in [1,N]$ is an arbitrary mixing index for the $i$-th datapoint and $f$ represents the decoding network. $p^*$, $p$ and $t$ are $L_2$-normalized. The CLIP loss with MixCo is defined as:
\begin{align}
    \mathcal{L}_{\text{BiMixCo}} = - \sum_{i=1}^N\left[\lambda_i\cdot\log\left(
        \frac{\exp\left(\frac{p_i^*\cdot t_i}{\tau}\right)}
             {\sum_{m=1}^{N}\exp\left(\frac{p_i^*\cdot t_m}{\tau}\right)}
    \right)+ (1 - \lambda_i)\cdot\log\left(
        \frac{\exp\left(\frac{p_i^*\cdot t_{k_i}}{\tau}\right)}
             {\sum_{m=1}^{N}\exp\left(\frac{p_i^*\cdot t_m}{\tau}\right)}
    \right)
    \right] \nonumber\\
    - \sum_{j=1}^N\left[\lambda_j\cdot\log\left(
        \frac{\exp\left(\frac{p_j^*\cdot t_j}{\tau}\right)}
             {\sum_{m=1}^{N}\exp\left(\frac{p_m^*\cdot t_j}{\tau}\right)}
    \right)+ \sum_{\{l\mid k_l=j\}}(1 - \lambda_l)\cdot\log\left(
        \frac{\exp\left(\frac{p_l^*\cdot t_j}{\tau}\right)}
             {\sum_{m=1}^{N}\exp\left(\frac{p_m^*\cdot t_j}{\tau}\right)}
    \right)
    \right],   
\end{align}
where $\tau$ is a temperature hyperparameter, and $N$ is the batch size.

\paragraph{SoftCLIP.}
The soft contrastive loss \citep{scotti2024reconstructing} takes the dot product of CLIP image embeddings within a batch to generate the soft labels.
The loss (we omit the bidirectional component for brevity) is calculated between CLIP-CLIP and Brain-CLIP matrices as:
\begin{align}
    \mathcal{L}_{\text{SoftCLIP}} = - \sum_{i=1}^N\sum_{j=1}^N\left[
    \frac{\exp\left(\frac{t_i\cdot t_j}{\tau}\right)}
             {\sum_{m=1}^{N}\exp\left(\frac{t_i\cdot t_m}{\tau}\right)}
    \cdot\log\left(
        \frac{\exp\left(\frac{p_i\cdot t_j}{\tau}\right)}
             {\sum_{m=1}^{N}\exp\left(\frac{p_i\cdot t_m}{\tau}\right)}
    \right)
    \right].  
    \label{eqn:softclip}
\end{align}

\section{The Visualized Image Retrieval Results}
\label{sec:retrieval results}

Fig. \ref{fig:vis} depicts the image retrieval results of our visual brain decoding model, which is trained on 167 subjects from our consolidated HCP dataset.
The results are extracted from the testing of Subject 1, which is an unseen subject to this model. 
We show the top three images that have the highest retrieval similarities to the target image seen by Subject 1.
We can see from Fig. \ref{fig:vis} that the retrieved images exhibit significant visual similarity to the target image.
For example, in the first row, the target and retrieved images are all facial images, while the first retrieved image is exactly the target image. In the fourth row, all the retrieved images have two persons, as in the target image. In the last two rows, the retrieved images share similar colors and tones to the target images. 
Nonetheless, in some cases, such as the second retrieved image in the fifth row and the third retrieved image in the sixth row, the retrieval results are not satisfactory.

\end{document}